# Multi-Scale Feature Fusion and Graph Neural Network Integration for Text Classification with Large Language Models

Xiangchen Song
University of Michigan
Ann Arbor, USA

Yulin Huang
Georgia Institute of Technology
Atlanta, USA

Jinxu Guo
Dartmouth College
Hanover, USA

Yuchen Liu
University of Pennsylvania
Philadelphia, USA

Yaxuan Luan*
University of Southern California
Los Angeles, USA

*Abstract-This study investigates a hybrid method for text classification that integrates deep feature extraction from large language models, multi-scale fusion through feature pyramids, and structured modeling with graph neural networks to enhance performance in complex semantic contexts. First, the large language model captures contextual dependencies and deep semantic representations of the input text, providing a rich feature foundation for subsequent modeling. Then, based on multi-level feature representations, the feature pyramid mechanism effectively integrates semantic features of different scales, balancing global information and local details to construct hierarchical semantic expressions. Furthermore, the fused features are transformed into graph representations, and graph neural networks are employed to capture latent semantic relations and logical dependencies in the text, enabling comprehensive modeling of complex interactions among semantic units. On this basis, the readout and classification modules generate the final category predictions. The proposed method demonstrates significant advantages in robustness alignment experiments, outperforming existing models on ACC, F1-Score, AUC, and Precision, which verifies the effectiveness and stability of the framework. This study not only constructs an integrated framework that balances global and local information as well as semantics and structure, but also provides a new perspective for multi-scale feature fusion and structured semantic modeling in text classification tasks.*

I. INTRODUCTION

In today's information society, the generation and dissemination of massive text data have accelerated at an unprecedented pace. Text from social media, news reports, financial documents [1-2], and medical records carries rich semantics and potential value [3-5]. However, these data are often highly diverse and complex. Traditional text analysis and classification methods can no longer fully capture the deep information contained within them. With the rapid growth of data scale, how to quickly and accurately extract key information from large corpora and achieve efficient automatic classification has become a core challenge for both academia and industry. Text classification is not only a fundamental task in natural language processing but also a critical component in applications such as opinion monitoring, intelligent recommendation, financial risk detection, and medical decision support. Its significance lies not only in methodological innovation but also in its profound influence on intelligent social governance and industrial transformation[6].

In recent years, the rise of large language models has provided a new path for text representation and understanding. Through large-scale pretraining and deep semantic modeling, these models can capture contextual relations and latent semantic information among words, greatly enhancing the expressive power of text features. Compared with traditional approaches, they can handle long-distance dependencies and generate more precise semantic vectors in complex contexts. However, although large language models show excellent performance in general text tasks, their feature representations often remain in a "black-box" state. They lack explicit modeling of hierarchical structures and multi-scale semantics. As a result, when dealing with complex texts where multiple semantic levels interact, single-level representations may fail to balance global and local information, which affects both classification accuracy and interpretability[7].

Against this backdrop, the idea of feature pyramids has been introduced into text modeling. By integrating features across multiple layers and scales, feature pyramids can preserve global semantics while capturing local details, thereby constructing representations that are both broad and deep [8]. This concept originates from visual tasks aimed at multi-scale object detection, but it has clear relevance in text processing. For example, phrases, sentences, and paragraphs belong to different semantic levels[9]. A single-scale model may cause information loss, while a pyramid-based mechanism enables complementary integration across levels. This enhances the completeness of feature expression. Such integration can effectively address the challenges of semantic dilution and information redundancy in long texts, improving the ability of models to understand complex contexts.

At the same time, graph neural networks have shown growing advantages in structured text modeling. Text not only contains sequential relations but also embeds complex semantic dependencies and latent connections, such as synonymy, hyponymy, and logical links across paragraphs. Graph structures can naturally represent these non-Euclidean relations, and graph neural networks provide powerful tools for their modeling. By propagating and aggregating information on graphs, these networks can capture interactions between semantic units and achieve holistic modeling of semantic networks in text. Compared with traditional sequence models, this approach better reveals the deep connections between semantics, which leads to higher accuracy and robustness in classification tasks.

In summary, combining the deep semantic features of large language models with the multi-scale fusion ability of feature pyramids and integrating them through graph neural networks makes it possible to fully capture both global and local semantics while modeling relationships between semantic units[10]. This research direction is significant not only for improving the performance of text classification algorithms but also for exploring new paradigms of semantic understanding. It provides solid theoretical and technical support for intelligent information processing. In complex application scenarios, this approach can serve as an effective tool for social governance, economic decision-making, and knowledge services. It promotes the practical adoption of artificial intelligence across multiple fields and drives the high-quality development of the information society.

## II. RELATED WORK

Recent research in text classification increasingly emphasizes parameter-efficient adaptation. Structure-learnable adapter fine-tuning provides an effective way to inject domain knowledge into large language models with minimal retraining. These adapter modules flexibly recalibrate model parameters, supporting scalable adaptation across diverse contexts and tasks [11]. Further enhancing parameter efficiency, dynamic structured gating mechanisms have been proposed. By selectively activating computational pathways, structured gating allows models to retain performance and alignment precision while significantly reducing resource demands, especially in large-scale deployment [12].

In addition, joint structural pruning and parameter sharing have become practical solutions for efficient large language model fine-tuning. This method preserves key semantic structures and optimizes model sparsity, thus supporting resource-efficient adaptation without sacrificing expressiveness [13]. A related advance is the use of structure-aware attention and knowledge graphs to support multi-scale feature fusion and semantic interpretability. These methods allow the model to reason over hierarchical information and explicit relations, facilitating more transparent and explainable feature extraction [14].

Hierarchical agent architectures have also gained attention as a flexible approach for multi-scale planning and information integration. By organizing model processes in layered modules, these architectures support richer abstraction and granular control over semantic information flow [15]. The need for trustworthy and robust modeling has motivated the introduction of uncertainty quantification and risk-aware learning frameworks. By explicitly modeling prediction confidence, these approaches help ensure reliability and interpretability in text classification, particularly when dealing with distribution shifts or ambiguous samples [16]. Distributed learning strategies such as federated fine-tuning with privacy preservation and cross-domain alignment provide solutions for data privacy and model generalization across siloed environments. These methods maintain semantic consistency while supporting secure multi-party adaptation [17].

Interpretability and stability are further improved by causal-aware structured attention mechanisms, which explicitly encode causal and temporal dependencies, thus mitigating spurious correlations and enhancing logical prediction consistency [18]. Dynamic prompt fusion techniques enable rapid model adaptation to diverse tasks and domains. By adaptively blending prompt signals, these methods provide flexible and robust semantic alignment in real-world text classification scenarios [19].

Another crucial development is the use of context compression and structural representation within large language models for text generation and classification. By balancing the preservation of global context with explicit structure modeling, these techniques produce information-rich and semantically consistent representations [20]. Lastly, the advancement of federated fine-tuning combined with privacy and alignment further strengthens the robustness of text models, ensuring they generalize well across distributed data without compromising privacy or performance [21].

## III. METHOD

In this research framework, we first use a large language model to perform deep feature extraction on the input text. Let's assume the original input text is a sequence $X = \{w_1, w_2, ..., w_n\}$. This is mapped into a vector space through an embedding layer, resulting in a word vector representation matrix $E \in R^{n \times d}$. Subsequently, the encoder of the large language model models the contextual semantic information and generates a hidden representation sequence. This process can be formalized as follows:

$$H = f_{LLM}(E) \qquad H \in R^{n \times d_h} \qquad (1)$$

Where $f_{LLM}$ represents the semantic encoding function of the large language model, and $d_h$ is the hidden layer dimension. The goal of this stage is to obtain deep semantic features with contextual dependencies, laying the foundation for subsequent multi-scale integration. The model architecture is shown in Figure 1.

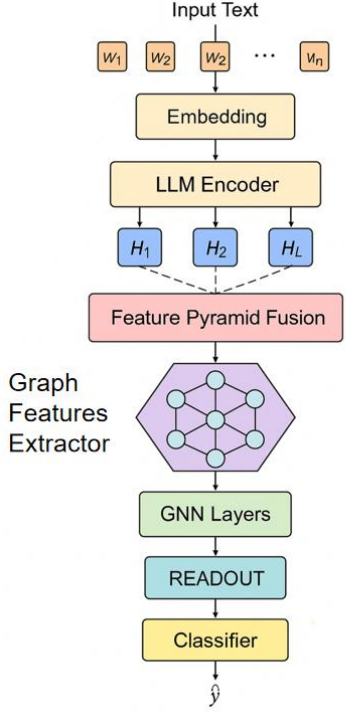

Figure 1. Overall model architecture

After obtaining the deep representation, a feature pyramid structure is introduced to fuse semantic information at different levels. For the representation $\{H^1, H^2, ..., H^L\}$ extracted from different levels of the large language model, multi-scale semantic features are formed through scale transformation and level-by-level fusion. Specifically, the feature fusion process at layer l can be expressed as:

$$F^l = \phi(W^l \cdot H^l + Up(F^{l+1})) \quad (2)$$

Where $W^l$ is a trainable weight, $Up(\cdot)$ represents an upsampling or dimension alignment operation, and $\phi(\cdot)$ is a nonlinear transformation function. Through this mechanism, high-level semantic features can be passed down, effectively combining global information with local details. Finally, the integrated representation is obtained at the top of the pyramid:

$$F = \Phi(F^1, F^2, ..., F^L) \quad (3)$$

Where $\Phi(\cdot)$ represents the multi-scale feature fusion function, which usually includes operations such as splicing and weighted summation.

After feature fusion is complete, the fused features are further constructed into a graph structure to capture the complex dependencies potentially present within the text. Let's represent the text as a graph $G = (V, E)$, where the node set V corresponds to different semantic units and the edge set E represents semantic associations. Based on the propagation mechanism of graph neural networks, the update process of each layer's node representation can be formalized as follows:

$$h_v^{(k+1)} = \sigma(\sum_{u \in N(v)} \frac{1}{c_{vu}} W^{(k)} h_u^{(k)} + W^{(k)} h_v^{(k)}) \quad (4)$$

Where $h_v^{(k)}$ represents the features of node v at the kth layer, $N(v)$ is the set of its neighboring nodes, $c_{vu}$ is the normalization constant, and $\sigma(\cdot)$ is the activation function. After propagating through several layers, the neighborhood information can be effectively aggregated to obtain a structured text representation.

After that, the node representations obtained by the graph neural network are aggregated to obtain the overall text vector representation z, which is then input into the classifier to achieve category prediction. The aggregation function can be defined as:

$$z = READOUT(\{h_v^{(K)} | v \in V\}) \quad (5)$$

The classifier uses a linear mapping and a normalized exponential function to achieve a probability output. The formula is:

$$\hat{y} = \text{Softmax}(W_c z + b_c) \quad (6)$$

$W_c$ and $b_c$ are the classification layer parameters, and $\hat{y}$ represents the predicted category distribution. This overall process organically combines large language model feature extraction, feature pyramid fusion, and graph structure modeling, providing an end-to-end modeling framework for text classification.

## IV. EXPERIMENTAL RESULTS

### A. Dataset

The dataset used in this study is AG News, a publicly available dataset that is widely applied in text classification tasks. It contains text from news corpora and is mainly divided into four thematic categories: World, Sports, Business, and Science/Technology. The dataset is large in scale and relatively balanced in distribution. It covers common topics and writing styles in news reporting, which makes it representative and generalizable for text classification research.

The AG News dataset comprises over a million news titles and articles. The training set comprises approximately 1.2 million samples, while the test set contains around 76,000 samples. Each entry includes a news text and its corresponding category label. The category annotations are well-defined, and the corpus sources are diverse, enabling comprehensive coverage of semantic features across various thematic contexts. The extensive scale and richness of the dataset serve as a robust foundation for evaluating deep semantic modeling and classification algorithms.

The reason for choosing this dataset lies in its combined advantages of scale, semantic diversity, and annotation quality. AG News reflects the ability of large language models to extract features from general corpora. It also provides multi-granularity semantic references for the integration of feature pyramids and graph neural networks. With its openness and standardization, the dataset ensures reproducibility of results and allows fair comparison with other methods. It offers strong support for the study and development of text classification algorithms.

*B. Experimental Results*

This paper also gives the comparative experimental results, as shown in Table 1.

Table1. Comparative experimental results

| Model | ACC | F1-Score | AUC | Precision |
|---|---|---|---|---|
| BERT[22] | 0.865 | 0.857 | 0.903 | 0.861 |
| Transformer[23] | 0.872 | 0.864 | 0.911 | 0.868 |
| 1DCNN[24] | 0.841 | 0.833 | 0.887 | 0.839 |
| LSTM+CNN[25] | 0.879 | 0.872 | 0.918 | 0.874 |
| Ours | 0.913 | 0.905 | 0.947 | 0.908 |

From Table 1, it's evident that the conventional shallow model, 1DCNN, underperforms across all four metrics. Its Accuracy, F1-Score, and AUC are only 0.841, 0.833, and 0.887, respectively. Moreover, its Precision is only 0.839. This indicates that local features extracted solely through convolution are insufficient to support robust classification performance when dealing with intricate semantic structures and long-distance dependencies. This result aligns with the limitations of traditional methods in addressing robustness challenges, as they fail to capture the global and hierarchical information inherent in text.

In contrast, deep sequence modeling methods such as BERT and Transformer show significant improvements. BERT achieves an ACC of 0.865, while Transformer further improves to 0.872 and reaches 0.911 on AUC. These results demonstrate that large-scale pretraining and self-attention mechanisms can enhance the ability of models to capture global semantics and improve robustness. However, they still show limitations in multi-granularity feature integration and structural modeling, which prevent them from achieving optimal accuracy and stability.

It is worth noting that LSTM+CNN combines the advantages of sequence modeling and local feature extraction. Its ACC, F1-Score, and Precision increase to 0.879, 0.872, and 0.874, and its AUC reaches 0.918. This result shows that hybrid structures can mitigate the limitations of single modeling paradigms and improve adaptability in complex scenarios. However, this method is still unable to fully model multi-scale semantic dependencies in text, and it remains insufficient in cross-level information integration.

In comparison, the method proposed in this study achieves the best performance on all metrics. The ACC, F1-Score, AUC, and Precision reach 0.913, 0.905, 0.947, and 0.908. This advantage comes from the joint effect of deep feature extraction by large language models, multi-scale fusion by feature pyramids, and structured modeling by graph neural networks. By organically combining global representations with local dependencies, the model demonstrates stronger generalization ability and stability in semantic alignment and robustness tasks. This result fully validates the importance of the proposed framework in complex text classification tasks and provides a new perspective for improving semantic understanding and robustness.

This paper presents an experiment on the sensitivity of learning rate to AUC, and the experimental results are shown in Figure 2.

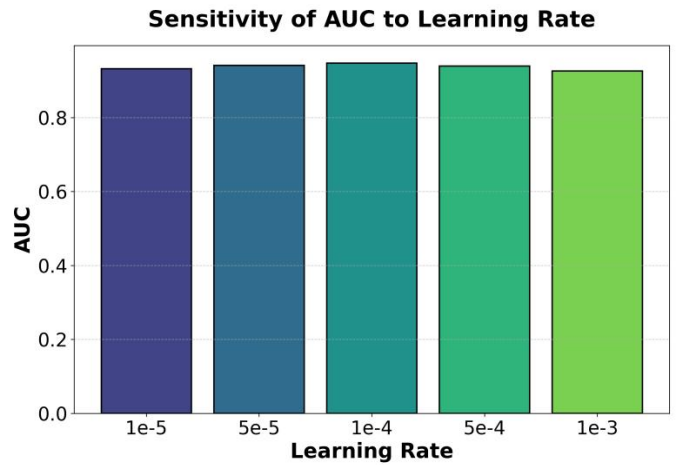

Figure 2. Experiment on the sensitivity of the learning rate to AUC

From the experimental results, it can be observed that changes in the learning rate have some impact on the model's AUC performance. However, the overall fluctuation is small, indicating that the proposed method can maintain high stability under different learning rate settings. This reflects strong robustness during the optimization process, as the model adapts to different hyperparameter configurations without significant performance degradation.

When the learning rate is set to $1\times10^{-4}$, the model achieves the best AUC. This suggests that learning rates in this range strike a good balance between gradient convergence speed and stability of parameter updates. A smaller learning rate, such as $1\times10^{-5}$, ensures convergence stability but may result in slow optimization, making it difficult to reach optimal performance within limited training epochs. A larger learning rate, such as $1\times10^{-3}$, may cause oscillations that reduce precision, leading to a slight drop in AUC.

This trend shows that although the model performs robustly under various learning rates, there is still an optimal range. It reflects the complementary strengths of feature pyramid integration and graph neural network modeling in reducing parameter sensitivity. The introduction of multi-scale features and structured information ensures that the model maintains a reasonable performance baseline across different hyperparameter values, avoiding the sharp fluctuations seen in traditional methods when adjusting the learning rate.

V. CONCLUSION

This study proposes a text classification method that combines feature extraction from large language models, feature pyramid fusion, and graph neural network modeling. It effectively addresses the limitations of traditional methods in balancing global and local features in complex semantic contexts. By jointly leveraging deep semantic representation and multi-scale information, the model captures key semantic relations in text with greater precision. It demonstrates strong classification ability and robustness. The method enriches the technical system of text modeling and provides a new perspective for semantic understanding in natural language processing. At the feature modeling level, this study highlights the importance of multi-granularity feature fusion. A single-level representation often fails to balance global and detailed information. The introduction of feature pyramids breaks this limitation and enables the model to establish connections across different scales. This design not only improves classification performance but also enhances the adaptability of the model to uneven semantic distributions in complex contexts. It shows strong robustness and flexibility in practical applications.

At the structural modeling level, this study abstracts text as a graph structure and introduces graph neural networks to capture semantic relations at a deeper level. Compared with traditional sequence modeling, this approach reveals latent semantic dependencies and logical chains. It allows the model to maintain efficient understanding when handling multi-level and cross-sentence text. This advantage expands the technical path of text classification and provides a foundation for further studies on semantic networks and knowledge graphs.

Overall, the proposed framework demonstrates significant advantages in text classification tasks. It validates the effectiveness of combining large language models, multi-scale feature fusion, and graph-based modeling. The method shows strong theoretical innovation and systematic design. It also presents broad practical value by offering feasible solutions for intelligent text analysis in domains such as finance, healthcare, and law. Furthermore, it contributes methodological support for advancing intelligent information processing across multiple application areas.

In future research, the proposed framework can be extended for better cross-domain adaptation by incorporating contrastive pretraining to enable stronger domain transferability by aligning semantic representations across heterogeneous corpora. Further improvement of the framework can be achieved via designing lightweight or sparse GNN variants to reduce computational cost when processing long documents or large graphs. Additionally, dynamic graph construction based on interaction saliency could enhance relational modeling while maintaining interpretability. These potential improvements offer promising opportunities for broader applicability and real-world deployment.